%% file: PaperForReview.tex
\documentclass[10pt,twocolumn,letterpaper]{article}

\usepackage{cvpr}          

\usepackage{graphicx}
\usepackage{amsmath}
\usepackage{amssymb}
\usepackage{booktabs}
\usepackage{fancyhdr}
\usepackage[pagebackref,breaklinks,colorlinks]{hyperref}

\usepackage{hyperref}
\usepackage{url}
\usepackage[accsupp]{axessibility}

\usepackage{graphicx}
\usepackage{amsmath}
\usepackage{amssymb}
\usepackage{booktabs}
\usepackage{times}
\usepackage{epsfig}
\usepackage{caption}
\usepackage{enumitem}
\usepackage{multirow}
\usepackage[misc]{ifsym}
\newcommand{\norm}[1]{\left\lVert#1\right\rVert}
\usepackage{makecell}
\usepackage{cases}
\usepackage{xcolor}
\usepackage{rotating}
\usepackage{chngpage}
\usepackage{amsmath}
\usepackage{subcaption}
\usepackage{appendix}

\usepackage[noend,ruled]{algorithm2e}
\usepackage{mathtools}

\newcommand\Mycomb[2][^b]{\prescript{#1\mkern-0.5mu}{}C_{#2}}

\usepackage[noend]{algpseudocode}
\newcommand*\Let[2]{\State #1 $\gets$ #2}

\usepackage{sidecap}
\sidecaptionvpos{figure}{t}

\usepackage[capitalize]{cleveref}
\crefname{section}{Sec.}{Secs.}
\Crefname{section}{Section}{Sections}
\Crefname{table}{Table}{Tables}
\crefname{table}{Tab.}{Tabs.}

\def\cvprPaperID{4418} 
\def\confName{CVPR}
\def\confYear{2023}

\begin{document}

\title{\Large STMT: A Spatial-Temporal Mesh Transformer for \\MoCap-Based Action Recognition}

\author{Xiaoyu Zhu$^1$, Po-Yao Huang$^2$$^\dag$, Junwei Liang$^3$$^\dag$, Celso M. de Melo$^4$, Alexander Hauptmann$^1$ \\
$^1$Carnegie Mellon University, $^2$FAIR, Meta AI, $^{3}$HKUST (Guangzhou), $^{4}$DEVCOM Army Research Laboratory \\
}

\vspace{1cm}
\twocolumn[\maketitle\vspace{-1em}\input{fig1-overview}\bigbreak]

\begin{abstract}
\vspace{-5mm}
    \input{0_abstract}
\end{abstract}
\vspace{-0.5cm}
\let\thefootnote\relax\footnotetext{\dag Equal Contribution.}


\input{1_intro}

\input{2_related}

\input{3_method}

\input{4_exp}

\input{5_conclusion}
\input{6_limitation}

{\small
\bibliographystyle{ieee_fullname}
\bibliography{egbib}
}

\end{document}

%% file: fig1-overview.tex
\begin{center}
\vspace{-5mm}
  \includegraphics[width=0.8\linewidth, height=5cm]{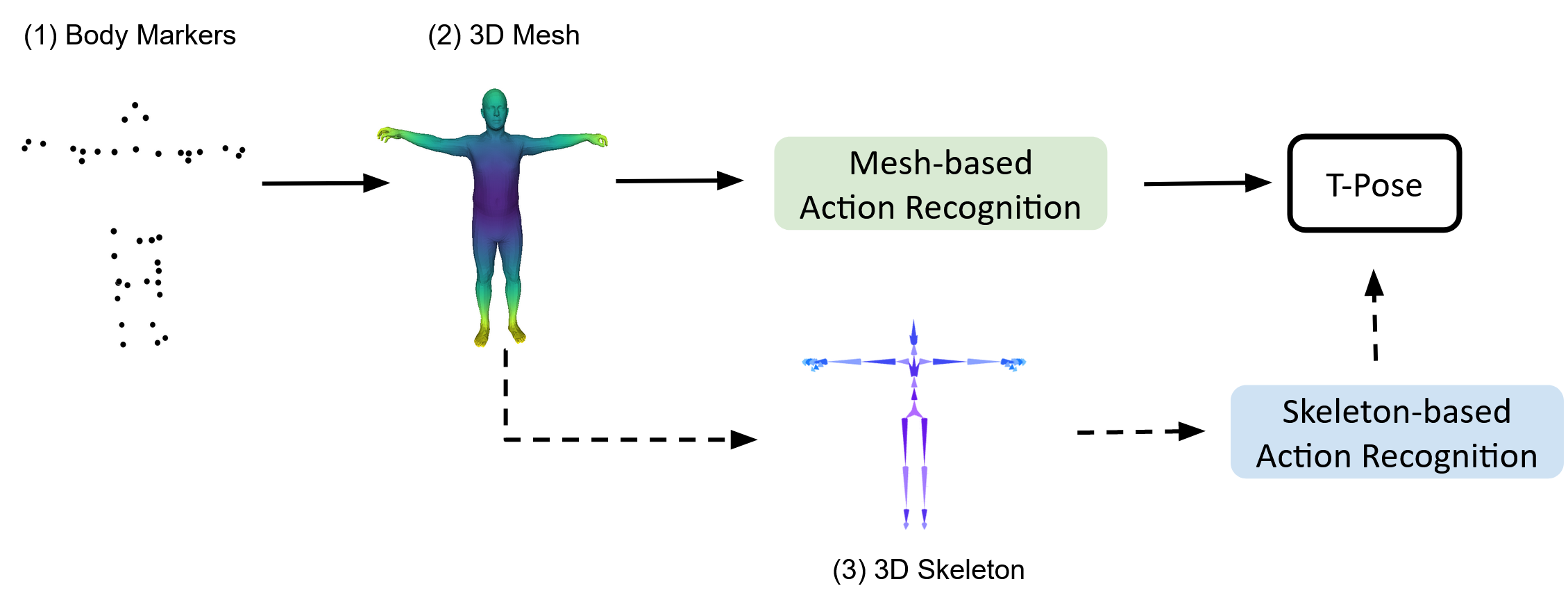}
\end{center}
\vspace{-2mm}
\captionof{figure}{Current state-of-the-art MoCap-based action recognition methods first convert body markers into a human body mesh, which is used to predict a standardized 3D skeleton. The 3D skeleton is used as input for action recognition models (dashed line). We propose a method that directly models the dynamics of raw mesh sequences (solid line). Our method saves the manual effort to derive skeleton representation, and achieves superior recognition performance by leveraging surface motion and body shape knowledge from meshes.}
\label{fig:fig1}

%% file: 0_abstract.tex
We study the problem of human action recognition using motion capture (MoCap) sequences. Unlike existing techniques that take multiple manual steps to derive standardized skeleton representations as model input, we propose a novel Spatial-Temporal Mesh Transformer (STMT) to directly model the mesh sequences. The model uses a hierarchical transformer with intra-frame off-set attention and inter-frame self-attention. The attention mechanism allows the model to freely attend between any two vertex patches to learn non-local relationships in the spatial-temporal domain. Masked vertex modeling and future frame prediction are used as two self-supervised tasks to fully activate the bi-directional and auto-regressive attention in our hierarchical transformer. The proposed method achieves state-of-the-art performance compared to skeleton-based and point-cloud-based models on common MoCap benchmarks. 
Code is available at {\url{https://github.com/zgzxy001/STMT}}.

%% file: 1_intro.tex
\fancypagestyle{firstpage}{
  \fancyhf{} 
  \renewcommand{\headrulewidth}{0pt} 
  \lfoot{} 
  \cfoot{1} 
  \rfoot{\textit{Approved for public release: distribution is unlimited.}} 
}
\thispagestyle{firstpage}

\section{Introduction}

Motion Capture (MoCap) is the process of digitally recording the human movement, which enables the fine-grained capture and analysis of human motions in 3D space \cite{AMASS:ICCV:2019,BABEL}. MoCap-based human perception serves as key elements for various research fields, such as action recognition \cite{ACCAD,SFU,MocapClub,EyesJapan,BABEL,07documentationmocap}, tracking \cite{07documentationmocap}, pose estimation \cite{achilles2016patient,VIBECVPR2020},  imitation learning \cite{zhao2012combining}, and motion synthesis \cite{07documentationmocap}. Besides, MoCap is one of the fundamental technologies to enhance human-robot interactions in various practical scenarios including hospitals and manufacturing environment \cite{hayes2017interpretable,kubota2019activity,zhu2011motion,malaise2018activity,Mehrizi2018ACV,menolotto2020motion}. For example, Hayes \cite{hayes2017interpretable} classified
automotive assembly activities using MoCap data of humans and objects.
Understanding human behaviors from MoCap data is fundamentally important for robotics perception, planning, and control. 

Skeleton representations are commonly used to model MoCap sequences.
Some early works \cite{BARNACHON2014238,5543273} directly used body markers and their connectivity relations to form a skeleton graph. However, the marker positions depend on each subject (person), which brings sample variances within each dataset. Moreover, different MoCap datasets usually have different numbers of body markers. For example, ACCAD \cite{ACCAD}, BioMotion\cite{Troje2002DecomposingBM}, Eyes Japan \cite{EyesJapan}, and KIT \cite{Mandery2015a} have 82, 41, 37, and 50 body markers respectively. This prevents the model to be trained and tested on a unified framework.
To use standard skeleton representations such as NTU RGB+D \cite{NTURGBD},  Punnakkal \emph{et al.}
 \cite{BABEL} first used Mosh++ to fit body markers into SMPL-H meshes, and then predicted a 25-joint skeleton \cite{liu2020ntu} from the mesh vertices \cite{MANO}.  Finally, a skeleton-based model \cite{Shi19} was used to perform action recognition. Although those methods achieved advanced performance, they have the following disadvantages. First, they require several manual steps to map the vertices from mesh to skeleton. Second, skeleton representations lose the information provided by original MoCap data (\emph{i.e.}, surface motion and body shape knowledge). To overcome those disadvantages, we propose a mesh-based action recognition method to directly model dynamic changes in raw mesh sequences, as illustrated in Figure~\ref{fig:fig1}.

Though mesh representations provide fine-grained body information, it is challenging to classify high-dimensional  mesh sequences into different actions. First, unlike structured 3D skeletons which have joint correspondence across frames, there is no vertex-level correspondence in meshes (\emph{i.e.}, the vertices are unordered). Therefore, the local connectivity of every single mesh can not be directly aggregated in the temporal dimension. Second, mesh representations encode local connectivity information, while action recognition requires global understanding in the whole spatial-temporal domain.

To overcome the aforementioned challenges, we propose a novel \underline{S}patial-\underline{T}emporal \underline{M}esh \underline{T}ransformer (\emph{STMT}). 
\emph{STMT} leverages mesh connectivity information to build patches at the frame level,
and uses a hierarchical transformer which can freely attend to any intra- and inter-frame patches to learn spatial-temporal associations.
The hierarchical attention mechanism allows the model to learn patch correlation across the entire sequence, and alleviate the requirement of explicit vertex correspondence. 
We further define two self-supervised learning tasks, namely masked vertex modeling and future frame prediction, to enhance the global interactions among vertex patches.
To reconstruct masked vertices of different body parts, the model needs to learn prior knowledge about the human body in the spatial dimension. To predict future frames, the model needs to understand meaningful surface movement in the temporal dimension. To this end, our hierarchical transformer pre-trained with those two objectives can further learn spatial-temporal 
context across entire frames, which is beneficial for the downstream action recognition task.

We evaluate our model on common MoCap benchmark datasets. Our method achieves state-of-the-art performance compared to skeleton-based and point-cloud-based models. The contributions of this paper are three-fold: 
 \begin{itemize}
	\item We introduce a new hierarchical transformer architecture, which jointly encodes intrinsic and extrinsic representations, along with intra- and inter-frame attention, for spatial-temporal mesh modeling.
	\item We design effective and efficient pretext tasks, namely masked vertex modeling and future frame prediction, to enable the model to learn from the spatial-temporal global context.
	\item Our model achieves superior performance compared to state-of-the-art point-cloud and skeleton models on common MoCap benchmarks.
\end{itemize}

%% file: 2_related.tex
\section{Related Work}
\textbf{Action Recognition from Depth and Point Cloud.}
3D action recognition models have achieved promising performance with depth~\cite{wang18,3dfcnn,xiao19,Sanchez20,Liu2020} and point clouds~\cite{qi2017pointnetplusplus,liu2019meteornet,3dv,pstnet}. 
Depth provides reliable 3D structural and geometric information which characterizes informative human actions.
In MVDI~\cite{xiao19}, dynamic images~\cite{bilen16} were extracted through multi-view projections from depth videos for 3D action recognition.
3D-FCNN~\cite{3dfcnn} directly exploited a 3D-CNN to model depth videos.
Another popular category of 3D human action recognition is based on 3D point clouds.
PointNet~\cite{qi2016pointnet} and PointNet++~\cite{qi2017pointnetplusplus} are the pioneering works contributing towards permutation invariance of 3D point sets for representing 3D geometric structures.
Along this avenue, MeteorNet~\cite{liu2019meteornet} stacked multi-frame point clouds and aggregates local features for action recognition.
3DV~\cite{3dv} transferred point cloud sequences into regular voxel sets to characterize 3D motion compactly via temporal rank pooling.
PSTNet~\cite{pstnet} disentangled space and time to alleviate point-wise spatial variance across time.
Action recognition has shown promising results with 3D skeletons and point clouds. Meshes, which are commonly used in representing human bodies and creating action sequences, have not been explored for the action recognition task. In this work, we propose the first mesh-based action recognition model. 
\\

\noindent\textbf{MoCap-Based Action Recognition}.
Motion-capture (MoCap) datasets \cite{ACCAD,SFU,MocapClub,EyesJapan,BABEL,07documentationmocap,celso_vision_based} serve as key elements for various research fields, such as action recognition \cite{ACCAD,SFU,MocapClub,EyesJapan,BABEL,07documentationmocap,celso_vision_based}, tracking \cite{07documentationmocap}, pose estimation \cite{achilles2016patient,VIBECVPR2020},  imitation learning \cite{zhao2012combining}, and motion synthesis \cite{07documentationmocap}.
MoCap-based action recognition was formulated as a skeleton-based action recognition problem \cite{BABEL}. 
Various architectures have been investigated to incorporate skeleton sequences.
In~\cite{du2015,zhang2017,Liu_2017_CVPR}, skeleton sequences were treated as time-series inputs to RNNs. 
\cite{hou2016,wang2016} respectively transformed skeleton sequences into spectral images and trajectory maps and then adopted CNNs for feature learning.
In~\cite{yan2018}, Yan \etal leveraged GCN to model joint dependencies that can be naturally represented with a graph. In this paper, we propose a novel method to directly model the dynamics of raw mesh sequences which can benefit from surface motion and body shape knowledge.
\\

\noindent\textbf{Masked Autoencoder.} Masked autoencoder has gained attention in Natural Language Processing and Computer Vision to learn effective representations using auto-encoding. Stacked denoising
autoencoders~\cite{VincentLLBM10} treated masks as a noise type and used denoising autoencoders  to denoise corrupted inputs. ViT~\cite{vit} proposed a self-supervised pre-training task to reconstruct masked tokens. More recently, BEiT~\cite{beit} proposed to learn visual representations by reconstructing the discrete tokens~\cite{dalle}. MAE~\cite{he2021masked} proposed a simple yet effective asymmetric framework for masked image modeling. In 3D point cloud analysis, Wang \emph{et al.} \cite{occo} chose to first generate partial point clouds by calculating occlusion from random camera viewpoints, and then completed occluded point clouds using autoencoding. Point-BERT \cite{yu2021pointbert} followed the success of BERT \cite{devlin} to predict the masked tokens learned from points. However, applying self-supervised learning to temporal 3D sequences (\emph{i.e.} point cloud, 3D skeleton) has not been fully explored. One probable reason is that self-supervised learning on high-dimensional 3D temporal sequences is computationally-expensive. In this work, we propose an effective and efficient self-supervised learning method based on masked vertex modeling and future frame prediction.

%% file: 3_method.tex
\section{Method}

%

\begin{figure*}[t]
\centering
\makebox[\textwidth][c]{\includegraphics[width=1\textwidth, height=6cm]{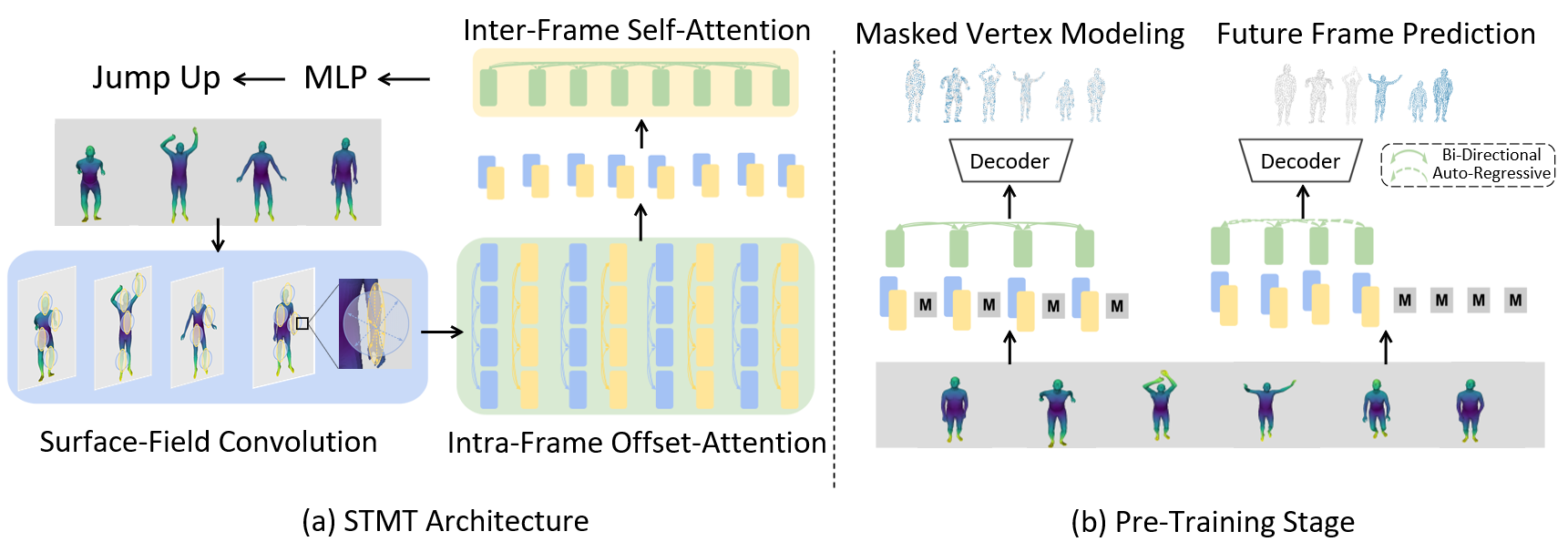}}%
\caption{Overview of the proposed framework. \textbf{(a) Overview of \emph{STMT}.} Given a mesh sequence, we first develop vertex patches by extracting both intrinsic (geodesic) and extrinsic (euclidean) features using surface field convolution. The intrinsic and extrinsic features are denoted by yellow and blue blocks respectively.  Those patches are used as input to the intra-frame offset-attention network to learn appearance features. Then we concatenate intrinsic patches and extrinsic patches of the same position. The concatenated vertex patches (green blocks) are fed into the inter-frame self-attention network to learn spatial-temporal correlations. Finally, the local and global features are mapped into action predictions by MLP layers. \textbf{(b) Overview of Pre-Training Stage.} We design two pretext tasks: masked vertex modeling and future frame prediction for global context learning. Bidirectional attention is used for the reconstruction of masked vertices. Auto-regressive attention is used for the future frame prediction task.
} 
\label{fig:stmt}
\vspace{-3mm}
\end{figure*}

\subsection{Overview}
\label{sub:overview}

In this section, we describe our model for mesh-based action recognition, which we call \emph{STMT}. The inputs of our model are temporal mesh sequences: $\mathbf{M} = ((\mathbf{P}_1, \mathbf{A}_1), (\mathbf{P}_2, \mathbf{A}_2), \cdots, (\mathbf{P}_t, \mathbf{A}_t))$, where \emph{t} is the frame number. $\mathbf{P}_i \in \mathbb{R}^{N \times 3}$ represents the vertex positions in Cartesian coordinates, where $N$ is the number of vertices.
 $\mathbf{A}_i \in \mathbb{R}^{N \times N}$ represents the adjacency matrix of the mesh. Element $\mathbf{A}_i^{mn} \in \mathbf{A}_i$ is one when there is an edge from vertex $V_m$ to vertex $V_n$, and zero when there is no edge. The mesh representation with vertices and their adjacent matrix is a unified format for various body models such as SMPL \cite{smpl2015}, SMPL-H \cite{MANO}, and SMPL-X \cite{SMPLX}. In this work, we use SMPL-H body models from AMASS \cite{AMASS:ICCV:2019} to obtain the mesh sequences, but our method can be easily adapted to other body models.

Mesh's local connectivity  provides fine-grained information. Previous methods \cite{hanocka2019meshcnn,diffusion} proved that explicitly using surface (\emph{e.g.}, mesh) connectivity information can achieve higher accuracy in shape classification and segmentation tasks. However, classifying temporal mesh sequences is a more challenging problem, as there is no vertex-level correspondence across frames. This prevents graph-based models from directly aggregating vertices in the temporal dimension. Therefore, we propose to first leverage mesh connectivity information to build patches at the frame level, then use a hierarchical transformer which can freely attend to any intra- and inter-frame patches to learn spatial-temporal associations. In summary, it has the following key components:

\begin{itemize}
    \item \textbf{Surface Field Convolution} to form local vertex patches by considering both intrinsic and extrinsic mesh representations.
    \item \textbf{Hierarchical Spatial-Temporal Transformer} to learn spatial-temporal correlations of vertex patches.
    \item \textbf{Self-Supervised Pre-Training} to learn the global context in terms of appearance and motion. 
    
\end{itemize}

See Figure~\ref{fig:stmt} for a high-level summary of the model, and
the sections below for more details.

\subsection{Surface Field Convolution}
\label{sub:surface}

Because displacements in grid data are regular, traditional convolutions can directly learn a kernel for elements within a region. However, mesh vertices are unordered and irregular. Considering the special mesh representations, we represent each vertex by encoding features from its neighbor vertices inspired by \cite{qi2016pointnet, qi2017pointnetplusplus}.
To fully utilize meshes' local connectivity information, we consider the mesh properties of extrinsic curvature of submanifolds and intrinsic curvature of the manifold itself. Extrinsic curvature between two vertices is approximated using Euclidean distance. Intrinsic curvature is approximated using Geodesic distance, which is defined as the shortest path between two vertices on mesh surfaces. We propose a light-weighted surface field convolution to build local patches, which can be denoted as:
\vspace{-3mm}
\begin{equation}\label{eq:fvg}
\boldsymbol{F}'^{(x,y,z)}_{VG} = \sum_{(\delta_x, \delta_y, \delta_z) \in G
(x,y,z)}\boldsymbol{W}^{(\delta_x, \delta_y, \delta_z)} \cdot \boldsymbol{F}^{(x+\delta_x, y+\delta_y, z+\delta_z)}  
\end{equation}
\begin{equation}\label{eq:fve}
\boldsymbol{F}'^{(x,y,z)}_{VE} = \sum_{(\zeta_x, \zeta_y, \zeta_z) \in E (x,y,z)}\boldsymbol{W}^{(\zeta_x, \zeta_y, \zeta_z)} \cdot \boldsymbol{F}^{(x+\zeta_x, y+\zeta_y, z+\zeta_z)} 
\end{equation}
\vspace{-2mm}

$G$ and $E$ is the local region around vertex $(x, y, z)$. In this paper, we use k-nearest-neighbor to sample  local vertices.  $(\delta_x, \delta_y, \delta_z)$ and $(\zeta_x, \zeta_y, \zeta_z)$ represent the spatial displacement in geodesic and euclidean space, respectively. $\boldsymbol{F}^{(x,y,z)}$ denotes the feature of the vertex at position $(x, y, z)$.

\subsection{Hierarchical Spatial-Temporal Transformer}

We propose a hierarchical transformer that consists of intra-frame and inter-frame attention. The basic idea behind our transformer is three-fold: (1) Intra-frame attention can encode connectivity information from the adjacency matrix, while such information can not be directly aggregated in the temporal domain because vertices are unordered. (2) Frame-level offset-attention can be used to mimic the Laplacian operator to learn effective spatial representations. (3) Inter-frame self-attention can learn feature correlations in the spatial-temporal domain.

\label{sub:hier}
\subsubsection{Intra-Frame Offset-Attention }

Graph convolution networks~\cite{gcn_2014} show the benefits of using a Laplacian matrix $\mathbf{L}=\mathbf{D}-\mathbf{E}$ to replace the adjacency matrix $\mathbf{E}$,
where $\mathbf{D}$ is the diagonal degree matrix. Inspired by this,
offset-attention has been proposed and achieved superior performance in point-cloud classification and segmentation tasks \cite{Guo_2021}. We adapt offset-attention to attend to vertex patches. 
Specifically, the offset-attention layer calculates the offset (difference) between the self-attention (SA) features
and the input features by element-wise subtraction. Offset-attention is denoted as:
\begin{align}\label{eq:oa}
    \boldsymbol{F}_{out} =OA(\boldsymbol{F}_{in})= & \mathbf{\phi}(\boldsymbol{F}_{in} - \boldsymbol{F}_{sa}) + \boldsymbol{F}_{in}.
\end{align}
where $\phi$ denotes a non-linear operator. $\boldsymbol{F}_{in} - \boldsymbol{F}_{sa}$ is proved to be analogous to discrete Laplacian operator \cite{Guo_2021}, \emph{i.e.} $\boldsymbol{F}_{in} - \boldsymbol{F}_{sa} \approx \boldsymbol{L}\boldsymbol{F}_{in}$. As Laplacian operators in geodesic and euclidean space are expected to be different, we propose to use separate transformers to model intrinsic patches and extrinsic patches. Specifically, the aggregated feature for vertex $V$ is denoted as:

\vspace{-3mm}
\begin{equation}\label{eq:M-ours}
\boldsymbol{F}'^{(x,y,z)}_V = OA_G(\boldsymbol{F}'^{(x,y,z)}_{VG}) \oplus OA_E(\boldsymbol{F}'^{(x,y,z)}_{VE})
\end{equation}

Here $F'^{(x,y,z)}_{VG} \in \mathbb{R}^{N\times d_g}$ and $F'^{(x,y,z)}_{VE} \in \mathbb{R}^{N\times d_e}$ are local patches learned using Equ.~\ref{eq:fvg} and Equ.~\ref{eq:fve}. $F'^{(x,y,z)}_{V} \in \mathbb{R}^{N\times d}$ denotes the local patch for position $(x,y,z)$, where $d = d_g+d_e$. The weights of $OA_G$ and $OA_E$ are not shared.

\subsubsection{Inter-Frame Self-Attention}

Given $F'_V$ which encodes local connectivity information, we use self-attention (SA) ~\cite{NIPS2017_3f5ee243} to learn {semantic affinities} between different vertex patches across frames.
Specifically, let
$\mathbf{Q,K,V}$ be the \emph{query, key} and \emph{value}, which are
generated {by applying linear transformations to the input features $F'_{V} \in \mathbb{R}^{N\times d}$} 
as follows:
\begin{align}\label{eq:qkv}
    (\boldsymbol{Q,K,V})           & = F'_{V} \cdot (\boldsymbol{W}_q, \boldsymbol{W}_k, \boldsymbol{W}_v)\nonumber         \\
    \boldsymbol{Q,K}               & \in \mathbb{R}^{N\times d_a},\quad\boldsymbol{V} \in \mathbb{R}^{N\times d} \nonumber \\
    \boldsymbol{W}_q, \boldsymbol{W}_k & \in \mathbb{R}^{d \times d_a},\quad \boldsymbol{W}_v \in \mathbb{R}^{d \times d}
\end{align}
where $\boldsymbol{W}_q$, $\boldsymbol{W}_k$ and $\boldsymbol{W}_v$ are the shared learnable {linear transformation, and $d_a$ is the dimension of the query and key vectors.} Then we can use the query and key matrices to {calculate the attention weights} via the matrix dot-product:
\vspace{-3mm}
\begin{equation}
    \boldsymbol{A}=(\Tilde{\alpha})_{i,j} =\mathrm{softmax} (\frac{\boldsymbol{Q} \cdot \boldsymbol{K}^{\mathrm{T}}}{\sqrt{d_a}}).
\end{equation}

\vspace{-2mm}
\begin{equation}\label{eq:sa}
    \boldsymbol{F}_{sa} = \boldsymbol{A}\cdot \boldsymbol{V}
\end{equation}

The self-attention output features $\boldsymbol{F}_{sa}$ are  the weighted sums of the value vector using the corresponding attention weights. Specifically, for a vertex patch in position $(x,y,z)$, its aggregated feature after inter-frame self-attention can be computed as: $\boldsymbol{F}_{sa}^{(x,y,z)} = \sum \boldsymbol{A}^{(x,y,z),(x',y',z')} \times \boldsymbol{V}^{(x',y',z')}$, where $(x',y',z') $ belongs to the Cartesian coordinates of $ \boldsymbol{F}_V'$.

\subsection{Self-Supervised Pre-Training}
\label{sub:self}

Self-supervised learning has achieved remarkable  results on large-scale image datasets \cite{he2021masked}. However, self-supervised learning
for temporal 3D sequences (\emph{i.e.} point cloud, 3D skeleton) remains to be challenging and has not been fully explored. There are two possible reasons: (1) self-supervised learning methods rely on large-scale datasets to learn meaningful patterns \cite{cole2022does}. However, existing MoCap benchmarks are relatively small compared to 2D datasets like ImageNet \cite{deng2009imagenet}. (2) Self-supervised learning for 3D data sequences is computationally expensive in terms of memory and speed. In this work, we first propose a simple and effective method to augment existing MoCap sequences, and then define two effective and efficient self-supervised learning tasks, namely masked vertex modeling and future frame prediction, which enable the model to learn global context. The work that is close to us is OcCO \cite{occo}, which proposed to use occluded point cloud reconstruction as the pretext task. OcCO has a computationally-expensive process to generate occlusions, including point cloud projection, occluded point calculation, and a mapping step to convert camera frames back to world frames. Different from OcCO, we randomly mask vertex patches or future frames on the fly, which saves the pre-processing step. Moreover, our pre-training method is designed for temporal mesh sequences and considers both bi-directional and auto-regressive attention.

\subsubsection{Data Augmentation through Joint Shuffle}
Considering the flexibility of SMPL-H representations, we propose a simple yet effective approach to augment SMPL-H sequences by shuffling body pose parameters. Specifically, we split SMPL-H pose parameters into five body parts: bone, left/right arm, and left/right leg. We use $I_{bone}, I_{leg}^{left}, I_{leg}^{right}, I_{arm}^{left}, I_{arm}^{right}$ to denote the SMPL-H pose indexes of the five body parts. Then we synthesize new sequences by randomly selecting body parts from five different sequences. We keep the temporal order for each part such that the merged action sequences have meaningful motion trajectories. Pseudocode for the joint shuffle is provided in Algorithm~\ref{algo}. The input to Joint Shuffle are SMPL-H pose parameters $\theta\in\mathbb{R}^{b \times t \times n \times 3}$, where $b$ is the sequence number, $t$ is the frame number, and $n$ is the joint number. We randomly select the shape $\beta$ and dynamic parameters $\phi$ from one of the five SMPL-H sequences to compose a new SMPL-H body model. Given $b$ SMPL-H sequences, we can synthesize $\Mycomb{5} = \frac{b!}{5!(b-5)!}\quad$ number of new sequences. We prove that the model can benefit from large-scale pre-training in Section~\ref{sub:analysis}.

\begin{algorithm}

  \caption{Pseudocode of STMT Joint Shuffle}\label{algorithm}
  \begin{algorithmic}[1]
  
    \Function{stmt\_joint\_shuffle}{$\theta\in\mathbb{R}^{b \times t \times n \times 3}, I_{bone}, I_{leg}^{left}, I_{leg}^{right}, I_{arm}^{left}, I_{arm}^{right}$}
    \Let{$\theta_s$}{$random\_sample(\theta, 5)$} \Comment{$\theta_s \in \mathbb{R}^{5 \times t \times n \times 3}$, randomly sample five SMPL-H sequences}
    
    \Let{$t_{max}$}{$get\_max\_length(\theta_s)$} \Comment{compute the maximum sequence length in $\theta_s$}
    \Let{$\theta_{new}$}{$Initialize(t_{max}, n, 3)$}
    \Let{$P$}{\{$I_{bone}, I_{leg}^{left}, I_{leg}^{right}, I_{arm}^{left}, I_{arm}^{right}$\}}

         \State \textbf{for} i in $0,1,2,3,4$ \textbf{do}
      \Let{\hspace{15px}$\theta_s$}{$repeat(\theta_s[i], (t_{max}, n, 3))$} \Comment{pad each sequence to the max length using repeating}
      \Let{\hspace{15px}$\theta_{new}$[$P[i]]$}{$\theta_s[i][P[i]]$} \Comment{assign the body-part sequence}
    
    \State \Return{$\theta_{new}$}
    \EndFunction
  \end{algorithmic}
  \label{algo}

\end{algorithm}

\subsubsection{Masked Vertex Modeling with Bi-Directional Attention} 
\label{subsub:mvm}

To fully activate the inter-frame bi-directional attention in the transformer, we design a self-supervised pretext task named Masked Vertex Modeling (MVM). The
model can learn human prior information in the spatial dimension by reconstructing masked vertices of different body parts. We randomly mask $r$ percentages of the input vertex patches, and force the model to reconstruct the full sequences. Moreover, we use bi-directional attention to learn correlations among all remaining local patches. Each patch will attend to all patches in the entire sequence. It models the joint distribution of vertex patches over the whole temporal sequences $x$ as the following product of conditional distributions, where $x_i$ is a single vertex patch:

\vspace{-3mm}
\begin{equation}
p(x) = \prod_{i=1}^{N} p(x_i | x_1,.., x_i,...,x_{N}) .
\label{eq:chain_rule}
\end{equation}
\vspace{-0.5mm}
Where $N$ is the number of patches in the entire sequence $x$ after masking. Every patch will attend to all patches in the entire sequence. In this way, bi-directional attention is fully-activated to learn spatial-temporal features that can accurately reconstruct completed mesh sequences.

\subsubsection{Future Frame Prediction with Auto-Regressive Attention}

The masked vertex modeling task is to reconstruct masked vertices in different body parts. The model can reconstruct completed mesh sequences if it captures the human body prior or can make a movement inference from nearby frames. As action recognition requires the model to  understand the global context, we propose the future frame prediction (FFP) task.
Specifically, we mask out all the future frames and force the transformer to predict the masked frames. 
Moreover, we propose to use auto-regressive attention for the future frame prediction task, inspired by language generation models like GPT-3 \cite{NEURIPS2020_1457c0d6}. However, directly using RNN-based models \cite{cho2014learning} in GPT-3 to predict future frames one by one is inefficient, as 3D mesh sequences are denser compared to language sequences. 
Therefore, we propose to reconstruct all future frames in a single forward pass. For auto-regressive attention, we model the joint distribution of vertex patches over a mesh sequence $x$ as the following product of conditional distributions, where $x_i$ is a single patch at frame $t_i$: 
\vspace{-2mm}
\begin{equation}
p(x) =  \prod_{i=1}^{N} p(x_i | x_1,x_2,...,x_M) .
\label{eq:chain_rule}
\vspace{-2mm}
\end{equation}

Where $N$ is the number of patches in the entire sequence $x$ after masking. $M = (t_i-1)\times n$, where $n$ is the number of patches in a single frame. Each vertex patch depends on all patches that are temporally before it. The auto-regressive attention enables the model to predict movement patterns and trajectories, which is beneficial for the downstream action recognition task.

\subsection{Training}
\label{sub:train}

In the pre-training stage, we use PCN \cite{8491026} as the decoder to reconstruct masked vertices and predict future frames. The decoder is shared for the two pretext tasks. Since
mesh vertices are unordered, the reconstruction loss and future prediction loss should be permutation-invariant. Therefore, we use Chamfer Distance (CD) as the loss function to measure the difference between the model predictions and ground truth mesh sequences. 

\vspace{-5mm}
\begin{equation}
\label{eq:cd}
\begin{split}
CD(M_{pred}, M_{gt}) = \frac{1}{|M_{pred}|}\sum_{x\in M_{pred}}\min_{y\in M_{gt}}\norm{x-y}_2 + \\ \frac{1}{|M_{gt}|}\sum_{y\in M_{gt}}\min_{x\in M_{pred}}\norm{y-x}_2 
\end{split}
\end{equation}

CD (\ref{eq:cd}) calculates the average closest euclidean distance between the predicted mesh sequences $M_{pred}$ and the ground truth sequences $M_{gt}$. The overall loss is a weighted sum of masked vertex reconstruction loss and future frame prediction loss:
\begin{equation}
\label{eq:pretrain_loss}
L = \lambda_1 CD(M_{pred}^{MVM}, M_{gt}) + \lambda_2 CD(M_{pred}^{FFP}, M_{gt})
\end{equation}

In the fine-tuning stage, we replace the PCN decoder with an MLP head. Cross-entropy loss is used for model training. 

%% file: 4_exp.tex
\section{Experiment}

\subsection{Datasets}

Following previous MoCap-based action recognition methods \cite{BABEL,sun2022locate}, we evaluate our model on the most widely used benchmarks: KIT\cite{Mandery2015a} and BABEL \cite{BABEL}. \textbf{KIT} is one of the largest MoCap datasets. It has 56 classes with 6,570 sequences in total. (2) \textbf{BABEL} is the largest 3D MoCap dataset that unifies 15 different datasets. BABEL has 43 hours of MoCap data
performed by over 346 subjects. We use the 60-class subset from BABEL, which contains 21,653 sequences  with single-class labels. We randomly split each dataset into training, test, and validation
set, with ratios of 70\%, 15\%, and 15\%, respectively. Note that existing action recognition datasets with skeletons only are not suitable for our experiments, as they do not provide full 3D surfaces or SMPL parameters to obtain the mesh representation.

\paragraph{Motion Representation.} Both KIT and BABEL's MoCap sequences are obtained from AMASS dataset in SMPL-H format. A MoCap sequence is an array of pose parameters over time, along with the shape and dynamic parameters. For skeleton-based action recognition, we follow previous work \cite{BABEL} which predicted the $25$-joint skeleton
from the vertices of the SMPL-H mesh.
The movement sequence is represented as $\mathbf{X} = (\mathbf{x}_1, \cdots, \mathbf{x}_L)$, where $\mathbf{x}_i \in \mathbb{R}^{J \times 3}$ represents the position of the $J$ joints in the skeleton in Cartesian coordinates. For point-cloud-based action recognition, we directly use the vertices of SMPL-H model as the model input. The point-cloud sequence is represented as $\mathbf{P} = (\mathbf{p}_1, \cdots, \mathbf{p}_L)$, where $\mathbf{p}_i \in \mathbb{R}^{V \times 3}$, and $V$ is the number of vertices. For mesh-based action recognition, we represent the motion as a series of mesh vertices and their adjacent matrix over time, as introduced in Section~\ref{sub:overview}. See Sup. Mat. for more details about datasets and pre-processing.

\begin{table*}[t]
    \vspace{-0.5cm}
    \begin{center}
    \normalsize{
        \addtolength{\tabcolsep}{-2pt}
        \begin{tabular}{l|c|cc|cc}
            \Xhline{1pt}
            \multirow{2}{*}{Method} & \multirow{2}{*}{Input} & \multicolumn{2}{c|}{KIT} & \multicolumn{2}{c}{BABEL-60} \\
           &  &  Top-1 (\%) & Top-5 (\%)& Top-1 (\%)& Top-5 (\%)\\
            \hline\hline
            2s-AGCN-FL \cite{2sagcn2019cvpr} (CVPR'19)          &   3D Skeleton        & 42.44 & 75.60 & 49.62 & 79.12  \\
            2s-AGCN-CE \cite{2sagcn2019cvpr} (CVPR'19)          &   3D Skeleton    & 57.46 & 81.54 & 63.57 & 86.77  \\
             CTR-GCN \cite{chen2021channel} (ICCV'21)          &   3D Skeleton         & 64.65 & 87.90 & 67.30 &  88.50 \\

            MS-G3D \cite{liu2020disentangling} (CVPR'20)          &   3D Skeleton          & 65.38 & 87.90 & 67.43 &  87.99 \\
            \hline
            PSTNet\cite{pstnet} (ICLR'21)         &   Point Cloud  & 56.93 &88.21  & 61.94 &   84.11\\
            SequentialPointNet\cite{seqpoint} (arXiv'21)            &   Point Cloud         & 59.75 &88.01  & 62.92 & 84.58  \\
            P4Transformer\cite{fan21p4transformer} (CVPR'21)           &   Point Cloud           & 62.15 & 88.01 & 63.54 & 86.55  \\
            \hline
            \textbf{STMT(Ours)}            &   Mesh            & \textbf{65.59} & \textbf{90.09} & \textbf{67.65} & \textbf{88.68}  \\

            \Xhline{1pt}
        \end{tabular}
    }
    \end{center}
    \vspace{-0.4cm}
        \caption{Experimental Results on KIT and BABEL Dataset. }
        
    \label{tab:sota}
    \vspace{-0.3cm}
\end{table*}

\subsection{Baseline Methods}
We compare our model with state-of-the-art 3D skeleton-based and point cloud-based action recognition models, as there is no existing literature on mesh-based action recognition. 2s-AGCN \cite{2sagcn2019cvpr}, CTR-GCN \cite{chen2021channel}, and MS-G3D \cite{liu2020disentangling} are used as skeleton-based baselines. Among those methods, 2s-AGCN trained with focal loss and cross-entropy loss are used as benchmark methods in the  BABEL dataset \cite{BABEL}. 
 For the comparison with point-cloud baselines, we choose PSTNet \cite{pstnet}, SequentialPointNet\cite{seqpoint}, and P4Transformer \cite{fan21p4transformer}. Those methods achieved top performance on common point-cloud-based action recognition benchmarks.

\subsection{Implementation Details}

For skeleton-based baselines, we use the official implementations of 2s-ACGN, CTR-GCN, and MS-G3D from \cite{2sACG}, \cite{ctr_git}, and \cite{ms_git}. 
 For point-cloud-based baselines, we use the official implementations of PSTNet, SequentialPointNet, P4Transformer from \cite{pst_git}, \cite{seq_git}, and \cite{p4d_git}. 
 We pre-train \emph{STMT} for 200 epochs with a batch size of 32. The model is fine-tuned for 50 epochs with a batch size of 64. Adam optimizer \cite{adam} is used with a learning rate of 0.0001 for both pre-training and fine-tuning. See Sup. Mat. for more implementation details.

\subsection{Main Results}
\label{sub:main_results}
\begin{figure}
\vspace{-4mm}
    \centering
    \includegraphics[width=9cm, height=6cm]{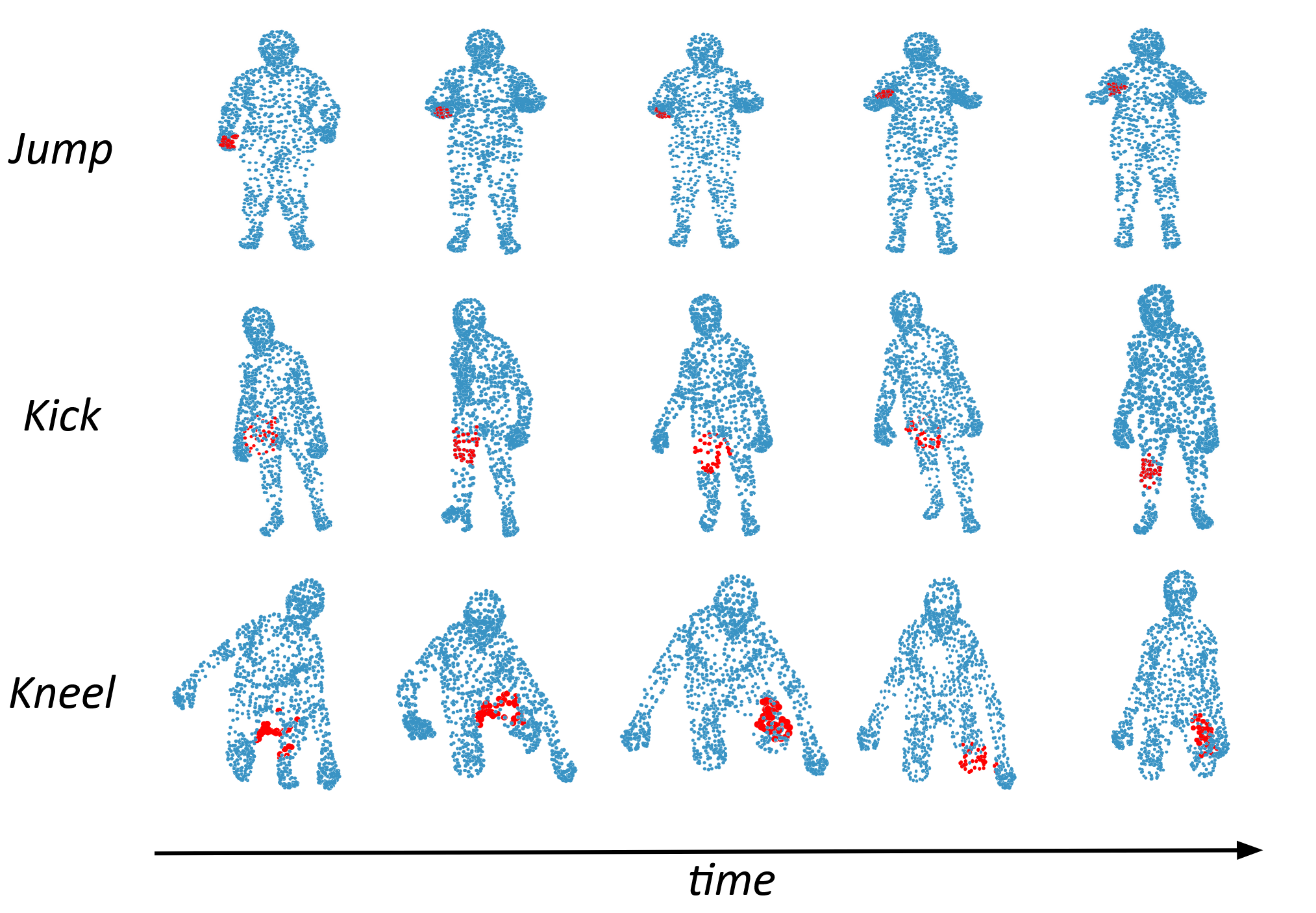}
      \vspace{-7mm}
    \captionof{figure}{Visualization of inter-frame attention. Red denotes the highest attention. 
    }
 
    \label{fig:attn}
   
  \vspace{-5mm}
\end{figure} 

\paragraph{Comparison with State-of-the-Art Methods.} As indicated in Table~\ref{tab:sota}, \emph{STMT} outperforms all other state-of-the-art models. Our model can outperform point-cloud-based models by 3.44\% and 4.11\% on KIT and BABEL datasets in terms of top-1 accuracy. Moreover, compared to skeleton-based methods which involve manual efforts to convert mesh vertices to skeleton representations, our model achieves better performance by directly modeling the dynamics of raw mesh sequences. 

We visualize the inter-frame attention weights of our hierarchical transformer in Figure~\ref{fig:attn}. We observe that the model can pay attention to key regions across frames. This supports the intuition that our hierarchical transformer can take the place of explicit vertex tracking by learning spatial-temporal correlations.

\begin{table}[!t]
\vspace{-2mm}
    \label{tab:aba}
      \centering
      \small
      \begin{tabular}{cccc|c}
\toprule
Intrinsic & Extrinsic & MVM & FFP & Top-1  (\%)\\
\midrule
\checkmark& & & & 63.40\\ 
\checkmark&\checkmark& & & 64.03 \\ 
\checkmark&\checkmark&\checkmark& &64.96\\
\checkmark&\checkmark&&\checkmark & 64.13 \\
\checkmark&\checkmark&\checkmark&\checkmark & \textbf{65.59} \\
\bottomrule
\end{tabular}
\caption{Performance of ablated versions. Intrinsic and Extrinsic stand for the intrinsic (geodesic) and extrinsic (euclidean) features in surface field convolution. MVM stands for Masked Vertex Modelling. FFP stands for Future Frame Prediction. 
}
\label{tab:aba}
\vspace{-0.4cm}
\end{table}
\subsection{Ablation Study}
 
\paragraph{Ablation Study of \emph{STMT}.} We test various ablations
of our model on the KIT dataset to substantiate our design decisions. We report the results in Table~\ref{tab:aba}. Note that Joint Shuffle is used in all of the self-supervised learning experiments (last three rows). We observe that each component of
our model gains consistent improvements. The comparison of the first two rows proves the effectiveness of encoding both intrinsic and extrinsic features in vertex patches. Comparing the last three rows with the second row, we observe a consistent improvement using self-supervised pre-training.   Moreover, the downstream task can achieve better performance with MVM compared to FFP. One probable reason is that the single task for future frame prediction is more challenging than masked vertex modeling, as the model can only see the person movement in the past. 
The model can achieve the best performance with both MVM and FFP, which demonstrates that the two self-supervised tasks are supplementary to each other.

\subsection{Analysis}
\label{sub:analysis}

\paragraph{Different Pre-Training Strategies.} We pre-train our model with different datasets and summarize the results in Table~\ref{tab:js}. The first row shows the case without pre-training. The second  shows the result for the model pre-trained on the KIT dataset (without Joint Shuffle augmentation). The third  shows the result for the model pre-trained on KIT dataset (with Joint Shuffle). We observe our model can achieve better performance with Joint Shuffle, as it can synthesize large-scale mesh sequences.

\begin{table}[!t]
\small
\centering
\vspace{-0.5cm}
\begin{tabular}{l|c}
\toprule
Method & Top-1 (\%) \\ \midrule
w/o pre-training  & 64.03\\
pre-training w/o JS  & 64.13 \\
pre-training w/ JS & \textbf{65.59} \\
\bottomrule
\end{tabular}
\caption{Comparison of Different Pre-Training Strategies. JS stands for Joint Shuffle.}
\label{tab:js}
  \end{table}
  
\begin{table}[t]
\vspace{-0.2cm}
\centering
\small
\begin{tabular}{c|c|c} 
\toprule
r  & Pre-Train Loss ($\times$ $10^4$) & Fine-Tune Accuracy (\%) \\
\midrule
0.1  &0.39 &64.44\\
0.3  & 0.41&64.55\\
\textbf{0.5}  & 0.40&\textbf{65.59}\\
0.7  & 0.43&64.19\\
0.9  & 0.48&65.07\\
Rand  & 0.43& 64.75\\
\bottomrule
\end{tabular}
\captionof{table}{Effect of Different Masking Ratios.}
\label{tab:ratio}
\vspace{-0.5cm}
 \end{table}

\paragraph{Different Masking Ratios.} We investigate the impact of different masking ratios. We report the converged pre-training loss and the fine-tuning top-1 classification accuracy on the test set in Table~\ref{tab:ratio}. 
We also experiment with the random masking ratio in the last row. 
For each forward pass, we randomly select one masking ratio from 0.1 to 0.9 with step 0.1 to mimic flexible masked token length. The model with a random masking ratio does not outperform the best model that is pre-trained using a single ratio (\emph{i.e.} 0.5). 
We observe that as the masking ratio increases, the pre-training loss mostly increases as the task becomes more challenging. However, a challenging self-supervised learning task does not necessarily lead to better performance. The model with a masking ratio of 0.7 and 0.9 have a high pre-train loss, while the fine-tune accuracy is not higher than the model with a 0.5 masking ratio. The conclusion is similar to the comparison of MVM and FFP training objectives, where a more challenging self-supervised learning task may not be optimal.

\begin{figure}
\vspace{-0.7cm}
    \centering
    \includegraphics[width=8cm, height=4cm]{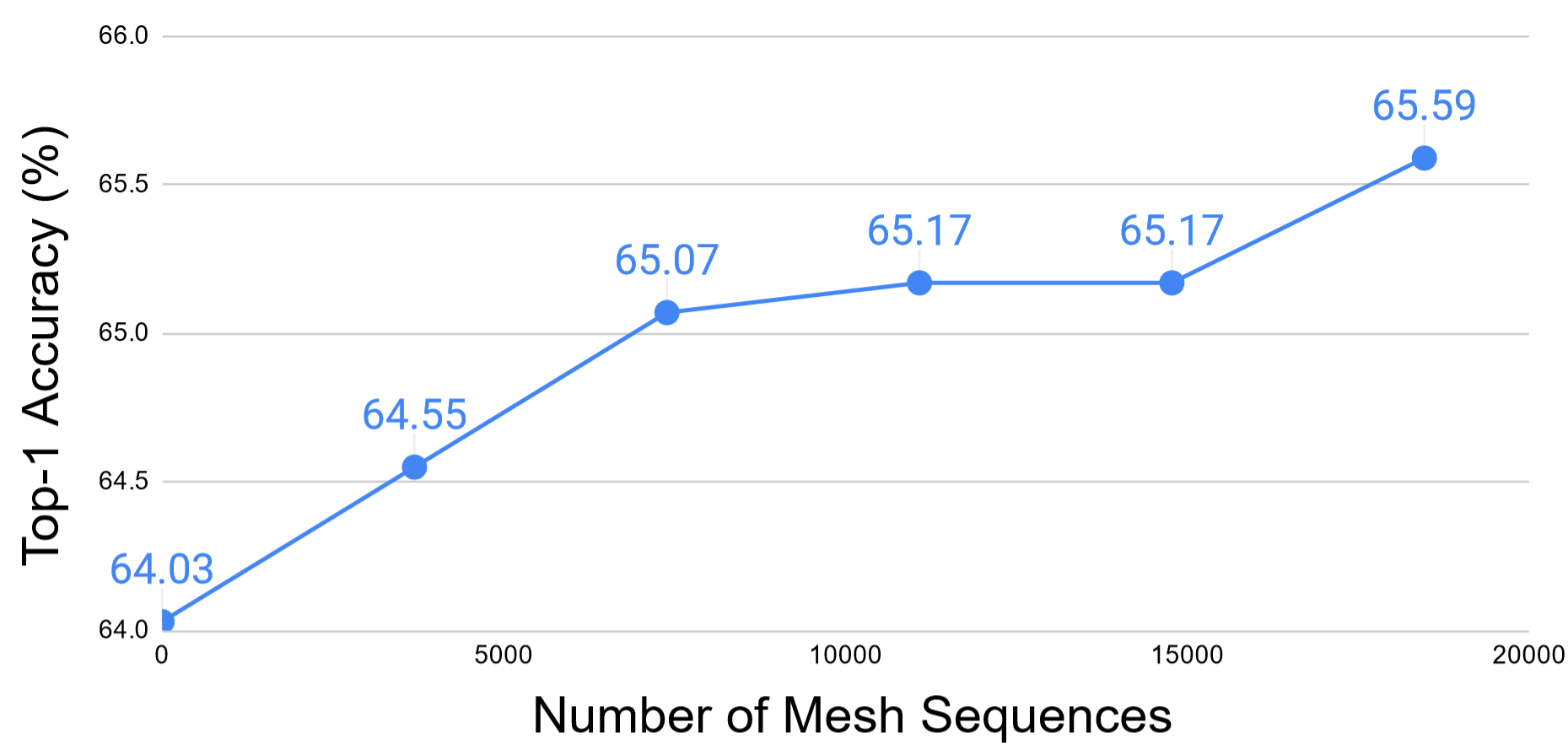}
    \vspace{-2mm}
    \captionof{figure}{Effect of Different Number of Mesh Sequences. 
    }
    \label{fig:meshnum}
   \vspace{-1mm}
\end{figure}

\paragraph{Different Number of Mesh Sequences for Pre-Training.} We test the effect of different numbers of mesh sequences used in pre-training.  We report the fine-tuning top-1 classification accuracy in Figure~\ref{fig:meshnum}. We observe that a large number of pre-training data can bring substantial performance improvement. The proposed Joint Shuffle method can greatly enlarge the dataset size without any manual cost, and has the potential to further improve model performance.

\paragraph{Experimental Results on Noisy Body Pose Estimations.} 
Body pose estimation has been a popular research field \cite{VIBECVPR2020,li2022cliff,jiang2022neuman}, but how to leverage the estimated 3D meshes for downstream perception tasks has not been fully explored. 
We apply the state-of-the-art body pose estimation model VIBE \cite{VIBECVPR2020} on videos of NTU RGB+D dataset to obtain 3D mesh sequences. Skeleton and point cloud representations are derived from the estimated meshes to train the baseline models (see Sup. Mat.). We report the results in Table~\ref{tab:ntu}. We observe that \emph{STMT} can outperform the best skeleton-based and point cloud-based action recognition model by 1.79\% and 3.44\% respectively. This shows that \emph{STMT} with meshes as input, is more robust to input noise compared to other state-of-the-art methods with 3D skeletons or point clouds as input. 

\begin{table}
\centering
\small
\begin{tabular}{c|c|c} 
\toprule
Method  & Input & Top-1 (\%) \\
\hline
\hline
2s-AGCN-FL \cite{2sagcn2019cvpr} & 3D Skeleton & 58.67\\
2s-AGCN-CE \cite{2sagcn2019cvpr} & 3D Skeleton& 57.49\\
CTR-GCN \cite{chen2021channel}& 3D Skeleton  &62.25 \\
MS-G3D \cite{liu2020disentangling} & 3D Skeleton & 60.01\\
\hline
PSTNet\cite{pstnet}  & Point Cloud & 51.48\\
SequentialPointNet\cite{seqpoint} & Point Cloud  & 60.60\\
P4Transformer\cite{fan21p4transformer}& Point Cloud  & 57.84\\
\hline
\textbf{STMT(Ours)} & Mesh & \textbf{64.04}  \\
\bottomrule
\end{tabular}
\captionof{table}{Experimental results on body poses estimated by VIBE \cite{VIBECVPR2020} on NTU RGB+D dataset. The skeleton, point cloud, and mesh representations are derived from the same noisy body estimations.}
\label{tab:ntu}
\vspace{-5mm}
\end{table}

%% file: 5_conclusion.tex
\section{Conclusion}
In this work, we propose a novel approach for MoCap-based action recognition. Unlike existing methods that rely on skeleton representation, our proposed model directly models the raw mesh sequences. Our method encodes both intrinsic and extrinsic features in vertex patches, and uses a hierarchical transformer to freely attend to any two vertex patches in the spatial and temporal domain. Moreover, two self-supervised learning tasks, namely Masked Vertex Modeling and Future Frame Prediction are proposed to enforce the model to learn global context.  Our experiments show that \emph{STMT} can outperform state-of-the-art skeleton-based and point-cloud-based models.

%% file: 6_limitation.tex
\section*{Acknowledgment}
This work was supported by the Army Research Laboratory and was accomplished under Cooperative Agreement Number W911NF-17-5-0003. 
This work used Bridges-2 GPU resources provided by PSC through allocation CIS220012 from the ACCESS program.